\documentclass{article}
\usepackage{ijcai22}

\usepackage{times}

\usepackage{soul}
\usepackage{hyperref}
\usepackage{url}
\usepackage[utf8]{inputenc}
\usepackage[small]{caption}
\usepackage{graphicx}
\usepackage{amsmath}
\usepackage{booktabs}
\usepackage[dvipsnames]{xcolor}
\usepackage{float}

\newcommand{\anonymous}[2]{#2}

\newcommand{\outline}[1]{}
\newcommand{\continuingissues}[1]{}
\newcommand{\issues}[1]{}

\newcommand{\bulletformat}[1]{\textsc{#1}}
\newcommand{\HumanFeedback}[0]{Human Feedback}
\newcommand{\humanfeedback}[0]{human feedback}

\newcommand{\KnowledgeBase}[0]{Knowledge Base}
\newcommand{\MachineLearner}[0]{Machine Learner}
\newcommand{\TeachingInterface}[0]{Teaching Interface}
\newcommand{\knowledgebase}[0]{\textsl{knowledge base}}
\newcommand{\machinelearner}[0]{\textsl{machine learner}}
\newcommand{\teachinginterface}[0]{\textsl{teaching interface}}

\newcommand{\ML}[0]{ML}
\newcommand{\LM}[0]{LM}
\newcommand{\NLP}[0]{NLP}
\newcommand{\MT}[0]{MT}

\newcommand{\VA}[0]{VA}
\newcommand{\ActiveLearningWithHumanCooperation}[0]{\textbf{AL+HC}}
\newcommand{\MTWithoutImportantWords}[0]{\textbf{MT-IW}}
\newcommand{\MTWithoutReplacements}[0]{\textbf{MT-VWR}}
\newcommand{\FullMT}[0]{\textbf{FullMT}}

\title{A Framework for Interactive Knowledge-Aided Machine Teaching}

\anonymous{
\author{
\affiliations
\textbf{Content Areas:} Intelligent User Interfaces, 
Knowledge Aided Learning, \\
Dialogue and Interactive Systems,
NLP Tools,
Cognitive Systems \\}
} {
\author{
Karan Taneja\footnote{Corresponding author}\and
Harshvardhan Sikka\And
Ashok Goel\\
\affiliations
Georgia Institute of Technology
\emails
\{karan.taneja, harshvardhan.sikka, ashok.goel\}@cc.gatech.edu
}
}

\begin{document}


\maketitle

\begin{abstract}

\anonymous{}{
\vspace{-2pt}
}

Machine Teaching (\MT) is an interactive process where humans train a machine learning model by playing the role of a teacher.
The process of designing an \MT~system involves decisions that can impact both efficiency of human teachers and performance of machine learners.
Previous research has proposed and evaluated specific \MT~systems but there is limited discussion on a general framework for designing them. 
We propose a framework for designing \MT~systems and also detail a system for the text classification problem as a specific instance.
Our framework focuses on three components i.e. teaching interface, machine learner, and knowledge base; and their relations describe how each component can benefit the others.
Our preliminary experiments show how MT systems can reduce both human teaching time and machine learner error rate.

\end{abstract}

\outline{\href{https://ijcai-22.org/calls-papers/}{Click for IJCAI 2022 Call For Paper}}

\anonymous{}{
\vspace{-9pt}
}

\section{Introduction}

\continuingissues{Acronyms}
\continuingissues{Ashok: Intro fits into one page}

\outline{MT definition}
Machine Teaching (\MT) is an emerging discipline 
that focuses on understanding and building interfaces for human teachers engaged in teaching machine learners.  
%
An MT system allows human teachers to teach machine learners using an interactive interface to build a machine learning (\ML) model. 
MT systems enables a wider community, beyond ML experts, to teach concepts to machine learners. 
According to \cite{Simard2017MachineSystems}, MT systems should have intuitive, efficient, and friendly interfaces that \textit{decouple} \MT~and \ML~processes such that the teaching does not require any knowledge of the underlying \ML~algorithms.
Another advantage of MT is the cost reduction in creating ML models while improving their performance.
\cite{Zhu2018AnTeaching,Liu2017IterativeTeaching,Zhu2015MachineEducation} studied \MT~as an optimization problem where learner performance needs to be maximized with minimum number of teaching examples. 
%
%
The optimization problem is given below where $\mathcal{D}$ is the dataset used for teaching, $\theta$ represents \ML~model parameters and $\eta$ is a scaling parameter.
\begin{equation}
\begin{aligned}
    \min_{\mathcal{D},\theta} \quad & \text{TeachingRisk} (\theta) + \eta \cdot \text{TeachingCost}(\mathcal{D})\\
\end{aligned}
\end{equation}

\begin{equation*}
\begin{aligned}
    \text{s.t.} \quad & \theta = \text{MachineLearning}(\mathcal{\mathcal{D}})\\
\end{aligned}
\end{equation*}
In general, teaching risk measures the learner error, such as empirical risk on a test set, while teaching cost measures the resources, such as number of examples, used in teaching. 
%
Since MT systems involve many components acting together and influencing the learning outcomes, there is a need to develop a framework that lays out design principles for ideating MT systems. 
%
%
Our proposed \MT~framework consists of three main components:
(i) \teachinginterface~which describes the machine state exposed to teacher and the \humanfeedback~that will be asked from the teacher,
(ii) \machinelearner~which describes the feedback interpretation and \ML~algorithm(s),
and (iii) \knowledgebase~which describes the domain knowledge and task-specific knowledge used by the system.
Our framework also proposes how these components will service each other and contribute towards the goals of machine teaching.

\outline{How we maximize performance / reduce teaching risk} 
As we shall describe in detail later, we aim to minimize teaching risk by (i) allowing granular feedback and combining it with knowledge, (ii) using an adaptive \machinelearner~which caters to online learning requirements and maximizes final performance, and (iii) prioritizing confusing examples using active learning.
%
%
We also aim to minimize teaching cost by (i) exposing the machine state to human teacher using ideas from interpretable \ML, (ii) using a \knowledgebase~to assist human teaching, and (iii) suggesting valuable examples for \humanfeedback.

\outline{Implementation setting: Intent classification problem in JW}
In this paper, we also describe implementation of an \MT~system for text classification. 
We particularly address the task of intent classification for 
\anonymous{an intelligent virtual assistant (\VA) that has been extensively}{Jill Watson (JW)}  deployed in real-world classroom settings \anonymous{[reference suppressed for blind review]}{\cite{Goel2019JillEducation}}.
An input to \anonymous{the \VA}{JW}~is classified into different intents for downstream processing in order to generate a response. 
For example, the question ``How do I turn in assignment 1?" will be classified as \textit{`submission'}, the question ``How much time will it take to solve the exam 2?" will be classified as \textit{`estimatedtime'}.
%
%
An important use case of our work are such systems where multiple \ML~models may need to be trained and deployed as part of a bigger application.
Each individual system can have an MT interface where a subject-matter expert can teach the machine to improve its performance using accumulated unlabeled examples over time.

\begin{figure*}[ht!]
    \centering
    \includegraphics[width=\linewidth]{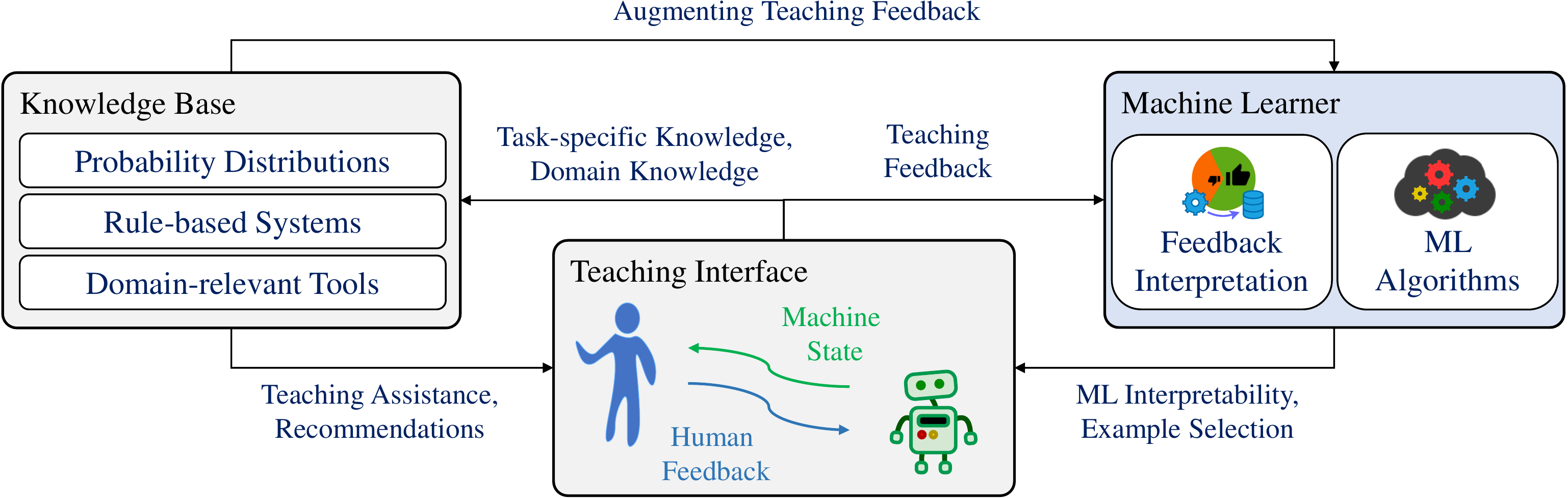}
    \caption{Three main components of proposed \MT~framework are \teachinginterface, \knowledgebase, and \machinelearner.
    \continuingissues{update figure with latest terminology?}
    }
    \label{fig:framework}
\end{figure*}

\outline{List of contributions} 
The main contributions of this paper are: 
(i) we formalize a framework for designing MT systems by laying out its components and their relations (Section \ref{sec:MTFramework}),
(ii) propose an MT system for text classification problem as a concrete instance of our framework (Section \ref{sec:MTimplementation}),
and (iii) we quantify the improvements obtained in our initial experiments in terms of performance and time efficiency (Section \ref{sec:results}). 
%

\section{Framework for MT System Design}
\label{sec:MTFramework}

\outline{What this section says}
In this section, we describe the main components of the proposed framework for designing \MT~systems (see Figure \ref{fig:framework}). 
The \teachinginterface~allows the teacher to interact with the \machinelearner~to train ML algorithms. This teaching process is supported by the \knowledgebase. 
Our aim is to keep the ML-expertise barrier low for human teachers while reducing the teaching cost and teaching risk.
We also discuss previous work in the context of our framework in Section \ref{sec:relatedwork}.
%

\subsection{\TeachingInterface} \label{subsec:interface}

There are two main parts of a \teachinginterface: Machine State and Human Feedback.

    \textbf{Machine State:} Machine state describes the information that is revealed to human teacher by the machine
    which helps the teacher in perceiving the \machinelearner's state.
    Interpretable \ML~methods \cite{Eiband2018BringingPractice,Das2020TaxonomyMethod} can be used to form a better understanding of the \machinelearner~for human teachers. 
    This allows teachers to adapt their teaching practices as the \machinelearner~improves. 
    
    \textbf{\HumanFeedback:} This describes the feedback that the teacher will provide to the machine. 
    \cite{Cui2021UnderstandingLearning} defined four broad types of feedback i.e. showing, categorizing, sorting and evaluating. 
    MT systems can employ one or more strategies to receive additional feedback on global as well as sub-input levels of granularity, such as token-level for input sentences or patch-level for images.

The \teachinginterface~aids (i) the \machinelearner~by providing data labels, granular feedback, etc., and (ii) the \knowledgebase~by building task-specific information as well as missing domain knowledge. 

\subsection{Machine Learner}

\outline{How a \machinelearner~is described}
A \machinelearner~is described by 
(i) the feedback interpretation mechanism used by the learner for exploiting feedback and 
(ii) the \ML~algorithms and related hyper-parameters.

\outline{Feedback interpretation aspects}
\textbf{Feedback Interpretation:} 
A \machinelearner~can accept feedback on different levels of granularity. 
Based on the feedback involved in teaching process, at least three feedback interpretation strategies that can be used are:

    \bulletformat{Input Features:} If features can be engineered using human feedback, the \machinelearner~can use them as inputs to the \ML~model \cite{Godbole2004DocumentLabels,Settles2011ClosingInstances,Jandot2016InteractiveClassification}.
    
    \bulletformat{Data Augmentation:} The feedback can be used for data augmentation which can increase both performance and robustness of \ML~models \cite{Rebuffi2021DataRobustness}.
    
    \bulletformat{Loss Function Augmentation:} The \humanfeedback~can be used to modify or augment loss functions. 
    For instance, \cite{Stiennon2020LearningFeedback} and \cite{Kreutzer2018CanFeedback} used human preferences as rewards in RL setting for 
    text summarization and 
    machine translation respectively. 
    \cite{He2016Human-in-the-LoopParsing} asked simple questions to non-experts for parsing task and use their answers for penalizing parser outputs.  

\outline{Learning algorithm aspects}
\textbf{\ML~Algorithms:} 
As the \machinelearner~receives a new teaching example, it needs to learn from it and present the new machine state to the teacher which requires online learning. 
The \machinelearner~can use a faster and smaller model for online tasks and a more powerful model for the end application.
Also, as more data is collected by the \machinelearner, it may need to adapt the underlying algorithm and/or related hyper-parameters to maximize its performance. 
The \machinelearner~can use techniques like neural architecture search \cite{Elsken2019NeuralSurvey} for this adaptive behavior. 

\outline{How \ML~helps human}
%
\label{subsubsec:ML>TI}

In addition to learning the task at hand, the \machinelearner~can aid human teaching in at least two ways:

    \bulletformat{\ML~Interpretability:} The \machinelearner~can provide its `interpretation' of an unlabeled example to expose \textit{Machine State} discussed in Section \ref{subsec:interface}. 
    \cite{Simard2017MachineSystems} reasoned that human teachers using MT systems should not need an understanding of the underlying \ML~algorithm(s). 
    We comply with this suggestion and also propose using model-agnostic interpretable ML for improving interaction with non-ML expert teachers. 
    
    \bulletformat{Example Selection:} 
    Example selection is the process of ranking unlabeled examples that are displayed to the human. 
    The machine can provide multiple unlabeled examples for feedback and rank them in order of decreasing confusion or increasing potential usefulness, such as in active learning \cite{Settles2009ActiveSurvey}. 
    Human teacher can use any example among these and provide feedback to the machine. 

\subsection{\KnowledgeBase}

The \knowledgebase~consists of domain-relevant information 
such as predefined probability distributions, rule-based systems, and any other tools.
Using pre-trained generative neural networks is one powerful way to define a probability distributions over the problem domain i.e. the input space.
Existing rule-based systems and other tools succinctly capture human expertise as procedural rules, relations, etc.
Our text classification MT system (described in Section \ref{sec:MTimplementation}) illustrates specific use cases of the \knowledgebase. 
In general, the \knowledgebase~can help MT system in two ways:

    \bulletformat{Aiding \TeachingInterface:} 
   The \teachinginterface~utilizes \knowledgebase~to assist humans in efficiently communicating their expertise to the machine. 
    For example, a word dictionary can be used to provide alternate word forms or synonyms that human teacher can use.
    In image domain, image inpainting models \cite{Bertalmio2000ImageInpainting} may recommend missing portions of an image.
    
    \bulletformat{Aiding \MachineLearner:} 
    Domain knowledge and task-specific knowledge provided by the \knowledgebase~can be used in augmenting human feedback for feedback interpretation by the \machinelearner.
    For example, data augmentation strategy can be supported by generative models or rule-based systems in text \cite{Kobayashi2018ContextualRelations,Wei2020EDA:Tasks}, images \cite{Frid-Adar2018GAN-basedClassification} and other domains. 
    


\section{Implementation of an MT System for Text Classification}
\label{sec:MTimplementation}

\begin{figure}[ht!]
    \centering
    \includegraphics[width=\linewidth]{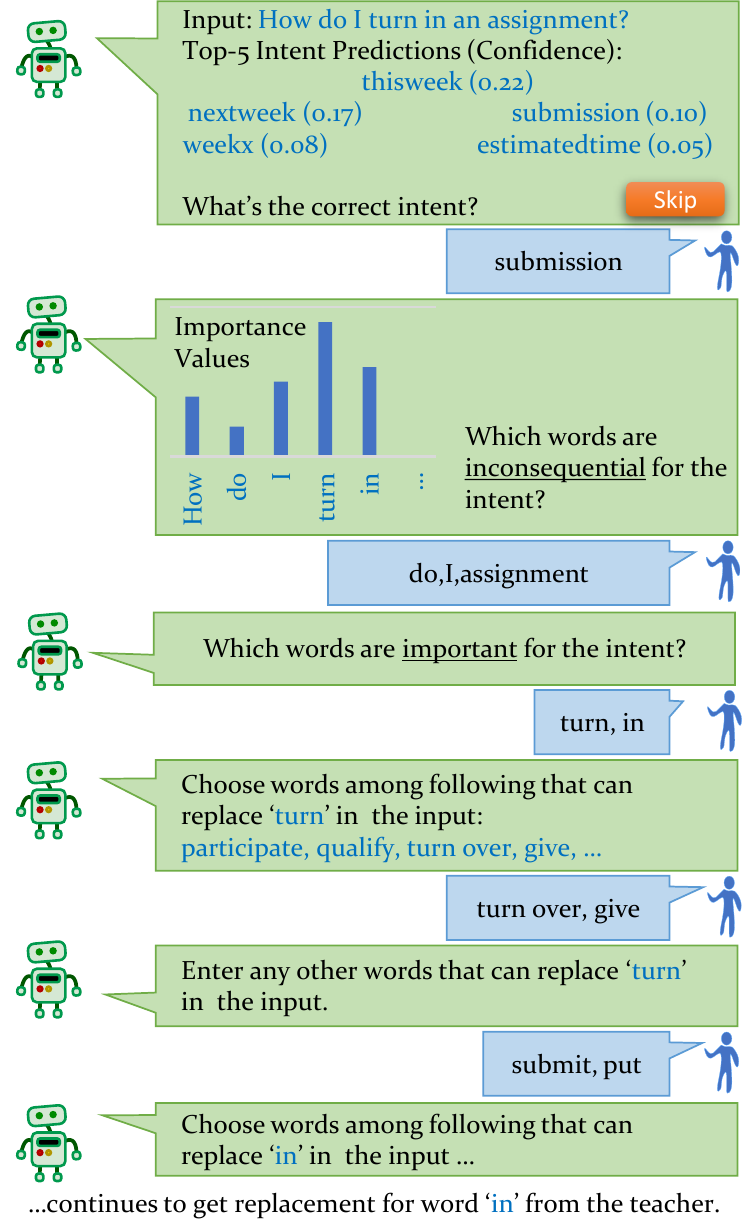}
    \caption{An example of interaction between our MT system implementation and human teacher.}
    \label{fig:implementation}
\end{figure}

\outline{What we do in this section}
In this section, we specify our implementation of an MT system which employs our proposed framework. 
The goal of this implementation is to create an interface for teaching a \machinelearner~about intent classification.
The learned intent classifier will classify incoming inputs to \anonymous{our \VA}{JW}~into one of 26 possible intents.
A sample interaction with our MT system implementation is given in Figure \ref{fig:implementation}.
In this example, input to \anonymous{the \VA~system}{JW} is ``How do I turn in an assignment?" and the correct intent is `\textit{submission}' 
since the input is inquiring about the submission process of an assignment, which is evident from the phrase `turn in' in the input.

\medskip \textbf{\TeachingInterface:} We built a command-line interface which has following two components:

    \bulletformat{\HumanFeedback:} The human teacher provides (i) a label for the novel unlabeled example, (ii) marks words that are important, (iii) marks words that are inconsequential in determining the intent and (iv) validates alternate word and phrase replacements for the important words. 
    In Figure \ref{fig:implementation}, teacher provides `submission' intent for the input, selects three words as inconsequential and two as important. 
    The replacements validated by the teacher for the word
    `turn' 
    include 
    `turn over', `give', `submit', and `put'.

    \bulletformat{Machine State:} Human teachers can judge machine state by observing (see example in Figure \ref{fig:implementation}):
    \begin{enumerate}
        \item \textbf{Model prediction on given unlabeled examples}: Before accepting an example for teaching, the teacher can see the top-$k$ predictions and corresponding confidence values. We used $k=5$ in our experiments.
        \vspace{-4pt}
        
        \item \textbf{Word importance values}: Before marking important and inconsequential words, the teacher can see importance values of words. (More details below.) 
        \vspace{-4pt}
        
        \item \textbf{Recommended replacements:} We use the \knowledgebase~(described later) to help the teacher in feeding alternate words and phrases for important words,   
    \end{enumerate}

\textit{How to calculate word importance values?:} 
Importance value of a word is determined by deleting it from the input and measuring the Kullback–Leibler divergence 
\cite{Kullback1951OnSufficiency} 
of the new output class distribution with the original distribution.
A higher divergence means that a word was more important for determining the class distribution. 
This is inspired from \cite{Ribeiro2016WhyClassifier} who measured token importance values by scoring them with many classifiers trained on input sentences where words were randomly removed.
Our online system uses the fast simplified procedure described above to calculate importance values.

\medskip \textbf{\MachineLearner~-- Feedback Interpretation:} 
We use the feedback from human teachers to augment data with two word replacement strategies.
Firstly, for words marked as important by teacher, we generate sentence variations by replacing them with all the replacements validated by human teacher.
For the example in Figure \ref{fig:implementation}, the word `turn' is replaced with `turn over', `give', `submit', and `put' to generate sentence variations.
Secondly, for words marked as inconsequential, we replace them with the top three replacements recommended by BERT masked language model (\LM) \cite{Devlin2019BERT:Understanding} residing in the knowledge base. This is done by hiding the word with a `[MASK]' token, feeding it as input to BERT masked LM and using the three output words with the highest confidence corresponding to `[MASK]' token. 

\textbf{\MachineLearner~-- Algorithms:}
We use bag-of-words perceptron model for (i) ranking unlabeled examples, and (ii) calculating word importance values displayed by \teachinginterface. 
The bag-of-words model trains in about 6-9 seconds and provides an online update since average labeling time (about 10 seconds) is more than the model update time.
We also run architecture search \cite{Elsken2019NeuralSurvey} in background to find better hyper-parameters based on the error on an evaluation set.
For the results reported in Section \ref{sec:results}, 
we fine-tune a pre-trained HuggingFace Tranformers model
\cite{Wolf2020HuggingFaceProcessing} 
with a classification head.
In particular, we used DistilBERT model 
\cite{Sanh2019DistilBERTLighter}
which is a light-weight language model ($40\%$ smaller than BERT).
%

\textit{How are teaching examples selected?} For example selection, we first reject the classes that have less than 1\% confidence and calculate Shannon entropy over remaining classes. 
The examples with the highest entropy are presented to human teacher and teacher is free to accept or skip the example for teaching. 
This simple technique requires minimal computation and time, even for large unlabeled sample pools.

\medskip \textbf{\KnowledgeBase:} 
In our implementation, the \knowledgebase~has three components:
%

    \bulletformat{BERT Masked \LM:} A word can be replaced with `[MASK]' token and fed to BERT masked \LM~\cite{Devlin2019BERT:Understanding} to get recommended replacements for assisting human teacher. 
    In Figure \ref{fig:implementation}, BERT provides some of the recommended replacements for word `turn', such as `participate' and `qualify'.
    The same model is also used for feedback interpretation as discussed earlier.

    \bulletformat{WordNet and Word Forms:} To recommend synonyms for validation by human teacher in \teachinginterface, we used 
    synonyms from \href{https://www.nltk.org/howto/wordnet.html}{WordNet} and 
    \href{https://github.com/gutfeeling/word_forms}{Python word\_forms package}. 
    Some replacements for `turn' in Figure \ref{fig:implementation} are recommended using this method, such as `turn over' and `give'.
    
    \bulletformat{Teacher Validated Replacements:} Word replacements validated by the teacher for important words are stored for future recommendations.
    In Figure \ref{fig:implementation}, validated replacements 
    `turn over', `give', `submit', and `put' 
    will be recommended first when the word `turn' is being replaced next time.



\begin{figure}[t]
    \centering
    \includegraphics[width=\linewidth]{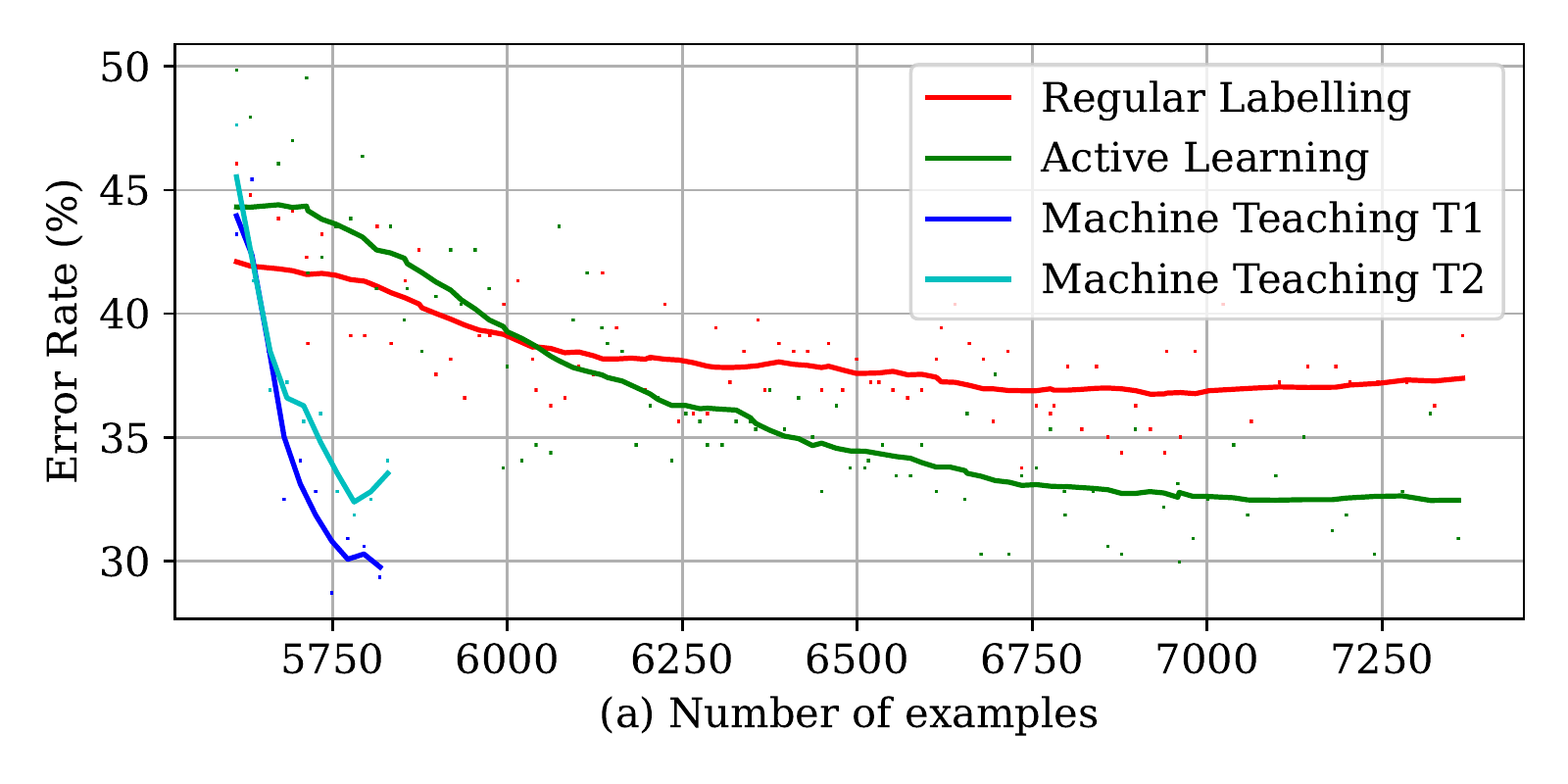} \includegraphics[width=\linewidth]{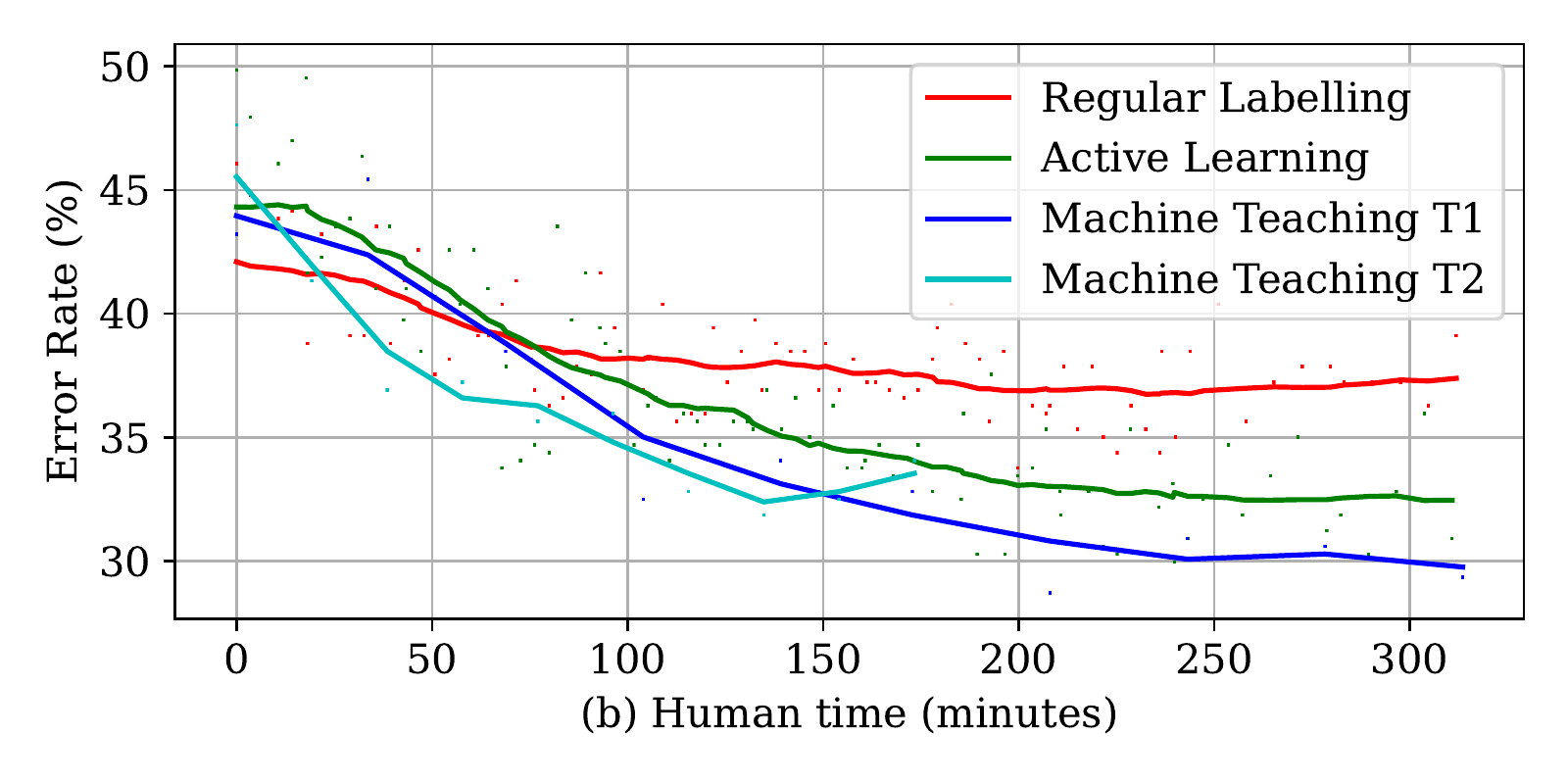}
    \caption{Comparing running-average classification error rate of our MT system with Regular Labeling and Active Learning with respect to number of examples and human time.
    }
    \label{fig:baseline_comparision}
\end{figure} 

\section{Experiments and Results}
\label{sec:results}

In this section, we describe our preliminary experiments and results. Our goal is to compare our MT implementation to regular labeling and active learning approaches. 

\textbf{Dataset:}
Our intent classification dataset has 28.1k examples 
which are generated by using 766 question templates labeled with intents.
\continuingissues{How were these templates collected?}
Each template contains placeholders for named entities. To generate examples, template placeholders are filled with a list of named entities relevant to problem domain.
We used 162 templates to generate 5.6k examples which are used to bootstrap the \machinelearner. 
Remaining 22.5k examples are used as novel examples. 
Some examples and more details can be found in supplementary materials.
Though this dataset is synthetic, it resembles use scenarios where users may be using similar inputs with different named entities.
For measuring the performance of our system, we use a test set with 317 manually labeled examples acquired from deployments of \anonymous{our \VA~system}{JW \cite{Goel2022AgentWatson}}. 

\textbf{Experimenter Bias:} For experiments with our MT system, human feedback was collected from the first two authors as independent teachers, denoted by T1 and T2, in a total of 315 and 174 minutes respectively. 
This may have introduced a favorable bias in results because they knew the final metrics to be collected and the technical details of the system.

\subsection{Comparison with Baselines}
We compare 
two 
baselines with our MT system:  

    \bulletformat{Regular Labeling:} Traditionally, large amounts of data is collected and annotated before developing \ML~models. 
    We simulate this data annotation process by including novel examples into the training dataset in \textit{random order} and measuring error rate as number of seen examples increase.
    
    \bulletformat{Active Learning:} In active learning, an \ML~algorithm scores the unlabeled data and constructs a query that includes most confusing examples. 
    A human annotator labels the examples in the query and the \ML~algorithm is trained by including these newly labeled examples. 
    We use the same active learning approach we used for example selection in our text classification MT system
    and simulate this data collection mechanism by including novel examples in the \textit{order of decreasing confusion} into the training dataset.
    %
    %

%
Figure \ref{fig:baseline_comparision} shows the running-average classification error rate of two baseline methods and our MT system with T1 and T2 users with increase in (a) the number of examples and (b) time. 
In Figure \ref{fig:baseline_comparision}a, we observe that active learning performs better than regular labeling (11.2\% relative decrease in error rate) as the number of examples are increased.
Our MT system leads to fastest reduction in error rate with only 205 teaching examples for T1 and 216 examples for T2. 
We observe a 4.2\% relative decrease in error rate compared to active learning and 14.9\% compared to regular labeling. 
We estimated human time spent in labeling examples by asking T1 and T2 to label 75 randomly selected examples  .
We found that additional feedback costs about 8$\times$ more time for T1 and 4$\times$ for T2. 
Comparing human time cost (Figure \ref{fig:baseline_comparision}b), we observe that our MT system achieves the largest reduction in error rate. 
In the initial part of labeling process, regular labeling shows lowest running-average error rate because it samples a more diverse set of examples from the novel examples
while active learning mechanism tends to give high confusion to similar examples. 
It is interesting that T2 spent almost half the time per example compared to T1 but achieved similar error rate improvements per unit time they spent (Figure \ref{fig:baseline_comparision}b). 
Rest of the paper only uses data collected from T1 for brevity. 

Each example taught using our MT system had a much higher impact on the classification performance than other two methods. 
We found that \humanfeedback~was used to create an average of 16 variations per example for T1. 
These are high-quality variations created by feedback interpretation mechanism using additional domain knowledge from BERT masked \LM.
We confirmed the quality of these variations by comparing with same amount of data generated using 
Easy Data Augmentation (EDA) technique \cite{Wei2020EDA:Tasks}
which involves random word insertions, swaps, deletions and synonym replacements. 
Using 16 variations per example generated using EDA, we observed a relative 11\% increase in error rate (3.15\% absolute error rate). 
We also note that we have not factored cognitive load in the teaching cost. 
Compared to our MT system, other baselines involve far more context switching since the teacher reviews 4-8$\times$ new examples in time they teach an example to the MT system.

\begin{figure}[t]
    \centering
    \includegraphics[width=\linewidth]{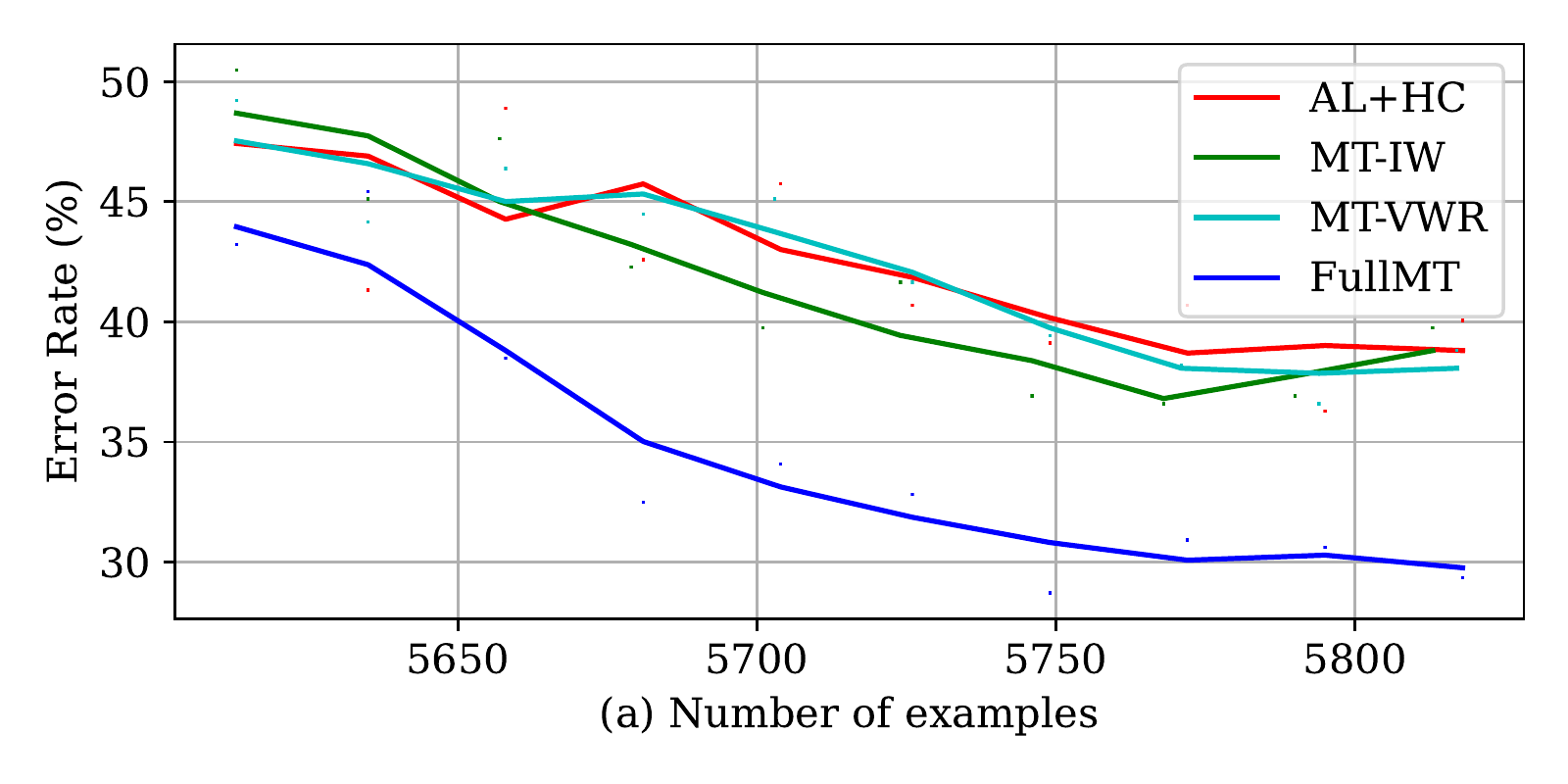} \includegraphics[width=\linewidth]{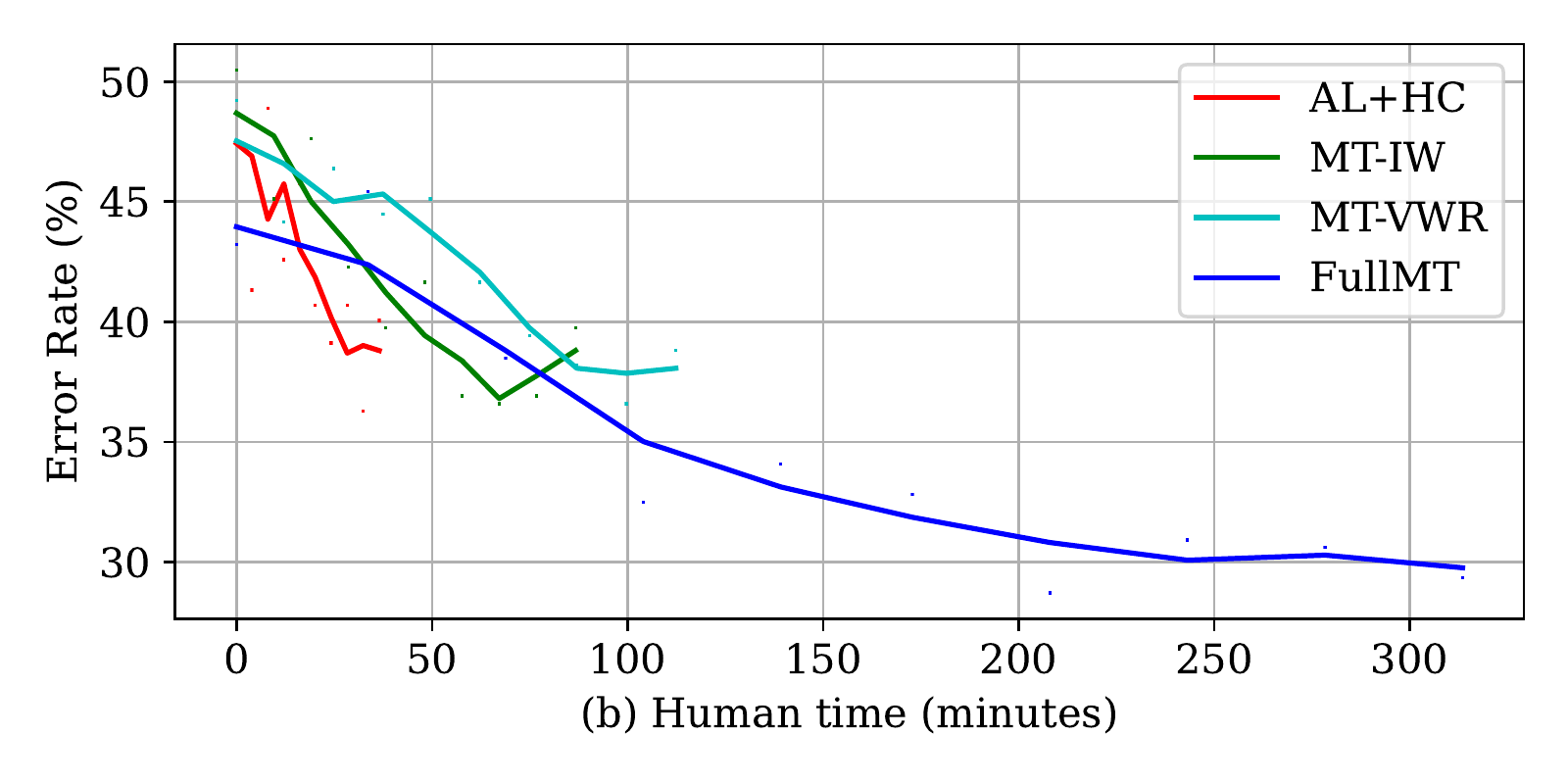}
    \caption{Comparing running-average classification error rate of our MT system with three ablated versions.}
    \label{fig:ablation}
\end{figure} 

\subsection{Ablation Studies}
Additional \humanfeedback~involves three parts: (i) highlighting inconsequential words that are not important in determining intent, (ii) highlighting important words that are important in determining intent and (iii) validating replacements for important words.
We observed that T1 spent 19\%, 9\% and 72\% of their time on above three tasks respectively. 
It took more time to highlight inconsequential words because
human teachers likely think about inconsequential and important words jointly, but highlight the inconsequential words first.
A very little time (about 1 second) was spent by T1 in skipping examples that were not deemed useful based on the example and intent predictions. 
On average, 1.16 examples were skipped for each accepted example.

In this subsection, we study each of these teaching sub-tasks and determine their importance using ablation experiments. We perform the following experiments:

\begin{itemize}
    \item \textit{Active Learning With Human Cooperation} (\ActiveLearningWithHumanCooperation): We use examples filtered by human teacher but do not use any additional feedback. 
    
    \item \textit{MT Without Important Words} \textbf{(\MTWithoutImportantWords)}: We use \humanfeedback~about inconsequential words and generate sentence variations by replacing them with words recommended by BERT masked LM.
    
    \item \textit{MT Without Validated Word Replacements} (\MTWithoutReplacements): We use feedback about both types of words. The incosequential words are replaced as in \MTWithoutImportantWords. The important words are replaced with intersection of BERT masked \LM~recommended words and WordNet synonyms.
    
    \item \textit{Full MT} (\FullMT): We use no ablations.
\end{itemize}

\outline{Explain ablation results}
The results for above four experiments are presented in Figure \ref{fig:ablation}. 
In terms of number of examples (Figure \ref{fig:ablation}a), we observe that using validated word replacements (\textbf{FullMT}) leads to highest reduction in the error rate. 
Using feedback about inconsequential words (\MTWithoutImportantWords) leads to a small improvement in performance over \ActiveLearningWithHumanCooperation. 
We find it interesting that \MTWithoutReplacements~leads to a small increase in the error rate over \MTWithoutImportantWords.
This suggests that dictionary synonyms are not necessarily good replacements for important words and human validated replacements are valuable for lowering error rate.

In terms of human time (Figure \ref{fig:ablation}b), \FullMT~is comparable to the ablated systems. 
We believe that this is because \FullMT~allows efficient injection of knowledge into \machinelearner.
It may also mean that spending more time on ablated versions may result in higher error rates (lower performance) because they lack injection of knowledge through validated replacements. 
To confirm this, we compared the best performing model generated by our MT system with same model trained on full dataset which has 28.1k examples, two orders of magnitude higher than number of teaching examples.
The \machinelearner~in our MT system outperforms the ML model trained using full data with a 0.94\% reduction in the absolute error rate. 
This is a small improvement, but it suggests that human teaching can inject knowledge that is not captured by the unseen data into the \machinelearner. 





\section{Related Literature}
\label{sec:relatedwork}

\continuingissues{past/present tense issue in related work}

\outline{Microsoft work in our framework}
\textbf{Machine Teaching:} 
\cite{Simard2014ICE:Problems} propose an interactive MT system for classification and information extraction.
They also use active learning to allow teachers to interactively select teaching examples from a pool of sorted unlabeled examples.
In addition to labels, human feedback also includes feature engineering.
%
%
In follow-up work, \cite{Ramos2020InteractiveModels} focus on creating interactive and intuitive interface for non-\ML~expert teachers to teach machine learners. 
For information extraction tasks, teachers can create a hierarchy of extracted entities which is an additional source of feedback.
To expose the machine state, they show errors in predictions on labeled examples in training set.
%
%
Our work proposes a general approach for designing MT systems. 
We aim for better understanding of (i) human feedback for machine, through feedback interpretation, and (ii) machine learner for human, through interpretable ML. 
We also focus on \knowledgebase~as a crucial component in assisting human teacher and augmenting teaching feedback.

\outline{UWisconsin and other theoretical work in our framework}
\cite{Zhu2018AnTeaching,Zhu2015MachineEducation} study MT as an optimization problem where teaching risk and teaching cost are minimized as discussed earlier.
\cite{Liu2017IterativeTeaching} theoretically study scenarios where teacher provides an example by synthesizing or selecting it from a pool to reduce risk at each step. 
Our framework allows creative ways in which the interactive process of teaching can be made more efficient for both teacher and learner. 
We also allow selection of difficult examples through active learning and promote usefulness of examples by allowing granular human feedback to machine learner. 

\outline{HIL-\ML~}
\textbf{Human-in-the-loop (HITL) \ML:} 
\cite{Cui2021UnderstandingLearning} surveyed HITL \ML~and how design choices affect interactive learning.
In their framework, they define four types of interaction: \textit{Showing}, \textit{Categorizing}, \textit{Sorting} and \textit{Evaluating}. 
Our current implementation uses \textit{Categorizing} at two levels: sentence-level for labelling and word-level for marking important and inconsequential words. 
It also uses \textit{Showing} in specifying alternate replacements for important words. 
In general, our framework allows elaborate feedback 
and more opportunities for feedback interpretation mechanisms.

\textbf{HITL Text Classification:}
Previous work has explored HITL text classification for interactive annotation and feature engineering.
\cite{Godbole2004DocumentLabels} created a user interface for text classification with active learning. Their interface allows human annotator to interactively engineer features for classifier and to see aggregate model statistics. 
\cite{Settles2011ClosingInstances} created a similar interface, additionally with an option to interactively label words with classes to use them as features. 
\cite{Simard2014ICE:Problems,Jandot2016InteractiveClassification} focus on interactive interfaces for feature creation by human annotators in addition to labeling examples. 
\cite{Wang2021PuttingSurvey} survey research in HITL natural language processing (\NLP).
Our framework is more general in terms of what feedback can be obtained from human teachers and how it can be used. 
In addition to this, we use \knowledgebase~in our framework for augmenting feedback and for assisting teachers.

\textbf{Data Augmentation:} 
Data augmentation improves model performance by augmenting training examples with perturbed copies. 
In \NLP, some common techniques include 
random word replacements, insertions, deletions, swaps \cite{Wei2020EDA:Tasks}, 
back-translating sentences to and back from a second language \cite{Sennrich2016ImprovingData}, 
using \LM s to generate sentence variations \cite{Kolomiyets2011Model-PortabilityAnalysis,Kobayashi2018ContextualRelations}. 
%
Our MT system implementation is an annotation tool relevant to early stages of \ML~development cycle and uses augmentation as a feedback interpretation mechanism.  
We use large pre-trained \LM s as part of the domain knowledge and incorporate human feedback into it for generating variations of labeled examples in a limited data setting. 
%

\textbf{Active Learning:} 
In active learning, an \ML~algorithm 
iteratively generates a query of examples 
for labelling by an oracle (such as human annotator) 
which leads to better performance with fewer training examples. 
%
For active learning with large unlabeled datasets in \NLP, pool-based sampling methods \cite{Settles2009ActiveSurvey,Schroder2020ANetworks} are used where queries are constructed by sampling a pool of unlabeled examples. 
%
Common methods rely on
model-based metrics like gradient magnitudes \cite{Settles2007Multiple-InstanceLearning} or
predictions-based metrics like entropy \cite{Hwa2004SampleParsing},
to select query from the sample pool. 
%
In our MT system, we use a na\"ive but fast entropy-based approach which helps in increasing learner's performance and reduces human time spent.



\section{Conclusion and Future Work}

We proposed a framework for MT system design with \teachinginterface, \machinelearner~and \knowledgebase~as three main components. 
Each component benefits other components in many ways to reduce teaching risk and teaching cost.
We presented a concrete example by implementing an MT system for text classification and discussed our results in a controlled experiment setting.
Our next goal is to build an intuitive and friendly web-based interface and conduct experiments with unbiased subjects and datasets in public domain. 
We also wish develop and test MT systems for other domains, such as image and speech. 

\continuingissues{Harsh's paper on Agent Smith \cite{Goel2022AgentWatson}}

\bibliographystyle{named}
\bibliography{ijcai22-multiauthor}



\section{Supplementary Material}
\subsection{Training Dataset}

Our training dataset for intent classification task is generated using templates for each intent. 
Some examples of intents and their templates are shown in Table \ref{tab:templates}. The name entities are inserted into placeholders marked by `\{object\}'. 
By iterating over a list of predefined entities for each placeholder, we generated a dataset for intent classification task. 

We have a total of 26 intents with 766 templates which generated about 28.1k examples. About 20\% of these templates were used to generate 5.6k examples for bootstrapping initial machine learner model and remaining 80\% were used to generate data for machine teaching process.

\begin{table}[ht!]
    \centering
    \begin{tabular}{|p{0.3\linewidth}|p{0.6\linewidth}|}
        \hline 
        
        \textbf{Intents} & \textbf{Example Template} \\ 
        
        \hline 
        submission & How do I submit the \{object\}? \\ 
        
        coursedescription & Will we learn about \{object\} in this class? \\ 
        
        teachingstaff & Who teaches this class?  \\ 
        
        officehours & When are office hours this week? \\ 
        
        lateworkpolicy & What is the penalty for submitting work past the deadline? \\ 
        
        
        importantdates & When is the \{object\}?  \\ 
        
        
        learning & What are the learning goals of this class? \\ 
        
        courseprerequisites & Do we need to know \{object\} to take this course? \\ 
        
        
        definition & Can you give an explanation for \{object\}? \\ 
        
        \hline
    \end{tabular}
    \caption{Examples of training templates used to generate training data.}
    \label{tab:templates}
\end{table}

\subsection{Future Research Directions}

Based on the machine state reported to the human teacher through the \teachinginterface, a teacher may choose different teaching strategies. 
An interesting future research direction would be to explore how different machine states affects human teaching and human performance. 
As an example, will revealing the error rate metric affect how teacher interacts with machine?
Also, does the teacher spend less time on an example when its confusion score is low? 

After understanding what different teaching behaviors imply, one can imagine building a mental model of human teacher in machine learner, leading to a mutual theory of mind. 
This mental model of teacher can be useful for machine learner in managing expectations and adapting its behavior. 

It will also be interesting to find if spending more time per example in \MT~improves human performance and the quality of data. Additionally, one can ask if lower context switching, as compared to regular labeling where new examples are presented far more frequently, leads to lower cognitive load. 


\end{document}